\documentclass[aps,pre,showpacs,amssymb,amsmath,nofootinbib]{revtex4}
\usepackage{graphicx}
\renewcommand{\b}[1]{\mathbf{#1}}
\begin{document}

\title{$L_{1}$ regularization is better than $L_{2}$ for learning and predicting chaotic systems}

\author{Zolt\'an Szab\'o}
\email[E-mail: ]{szzoli@cs.elte.hu}
\author{Andr\'as L{\H o}rincz}
\thanks{Corresponding author}
\email[e-mail: ]{lorincz@inf.elte.hu} \homepage[URL:
]{http://people.inf.elte.hu/lorincz}
\affiliation{Department of Information Systems\\
E\"{o}tv\"{o}s Lor\'{a}nd University\\
Budapest\\
Hungary H-1117\\}

\date{\today}

\begin{abstract}
Emergent behaviors are in the focus of recent research interest. It is then of
considerable importance to investigate what optimizations suit the learning and
prediction of chaotic systems, the putative candidates for emergence. We have
compared $L_{1}$ and $L_{2}$ regularizations on predicting chaotic time series
using linear recurrent neural networks. The internal representation and the
weights of the networks were optimized in a unifying framework. Computational
tests on different problems indicate considerable advantages for the $L_{1}$
regularization: It had considerably better learning time and better
interpolating capabilities. We shall argue that optimization viewed as a
maximum likelihood estimation justifies our results, because $L_{1}$
regularization fits heavy-tailed distributions -- an apparently general feature
of emergent systems -- better.
\end{abstract}


\maketitle

\section{Introduction}
We are interested in learning, representing, and predicting emergent phenomena.
The general view is that emergent dynamics brought about by nature (see, e.g.,
\cite{bak96how,Holland98Emergence} and references therein) and also, in some
man made constructs (see, e.g.,
\cite{Jensen98Self,Jonhson01Emergence,albert02statistical,Lodding04hitchhiker}
and references therein) show chaotic behavior and exhibit heavy-tailed
distributions. In turn, we are interested in learning, representing, and
predicting chaotic processes. Such processes could be of considerable
importance in neurobiology. According to the common view, the brain may apply
chaotic systems to represent and to control nonlinear dynamics
\cite{korn03isthere,gauthier03controlling}. Our study concerns the
identification of chaotic dynamics using recurrent linear neural networks, a
particular family of artificial neural networks (ANN).

Considering ANNs, sparsified representations have been of considerable research
interest and have shown reasonable success in representing natural scenes
\cite{sanger89contribution,levin94fast,deco95unsupervised,reed95similarities,kamimura97information,Olshausen97Sparse,Olshausen96Emergence,koene99discriminant}.
The underlying concept is to choose a representation, which can derive
(generate, reconstruct) the input by using the least number of components, that
is, which compresses information efficiently. From the theoretical point of
view, this assumption may be seen as a variant of Occam's razor principle (see,
e.g., \cite{chen02different} and references therein), provided that each
component has identical a priori probabilities. For a recent review on the
neurobiological relevance of sparse representations, see
\cite{olshausen04sparse}.

Numerical studies indicate that creation and pruning of connectivity matrices
of ANNs can be advantageous: it may decrease learning time and may increase
generalization capabilities (see, e.g.,
\cite{hinton89connectionist,williams94bayesian,hyvarinen00sparse}). Encouraging
experimental studies of joined representational and weight sparsifications have
also been undertaken in generative networks \cite{szatmary02nonnegative}. For a
thorough review on weight pruning methods, see \cite{haykin99} and references
therein. Weight sparsification is connected to structural risk minimization
\cite{Vapnik98Statistical}, a method that provides a measure for generalization
versus overfitting. Structural risk minimization is also related to certain
regularization methods (see, e.g.,
\cite{Vapnik98Statistical,evgeniou00regularization,chen02different} and
references therein).

We intend to demonstrate the advantages of sparsification. For the sake of
simplicity we shall deal with linear systems. This simplifications allowed us
to make a joined framework, which can treat sparse neural activities and sparse
neural connectivity sets on equal footing. The framework was constructed to
compare the performances of $L_{1}$ and $L_{2}$ norms. The chosen `battlefield'
is the identification of chaotic time series. This highly non-linear problem
should be of real challenge to the linear schemes. In turn, we also ask, how
good identification can be achieved using linear approaches.

The paper is organized as follows: Basic concepts and the recurrent neuronal
network to be studied are introduced in Section \ref{sec:Preliminaries}. The
mathematical formalism and the approximations are described in Section
\ref{sec:Methods}. Section \ref{sec:Experiments} reviews some numerical
experiments. Connections to information theory is discussed in Section
\ref{sec:Discussion}. Conclusions are drawn in the last section.

\section{\label{sec:Preliminaries}Preliminaries}
\subsection{Network architecture}
Neural architectures with reconstruction abilities shall be considered. In
particular, our formalism concerns Elman type recurrent networks, which belong
to the family of recurrent neural networks (RNNs). For a review on RNNs, see,
e.g., \cite{Tsoi97Discrete} and references therein.

The Elman network has $d_{\b{x}}$ input neurons, $d_{\b{u}}$ hidden neurons and
$d_{\b{o}}$ output neurons. The state, or activity, of each neuron is
represented by a real number in all layers. Activities of input, hidden, and
output neurons are indexed by time: quantities in the $t^{th}$ time instant are
$\b{x}_{t}\in \mathbb{R}^{d_{\b{x}}}$, $\b{u}_{t}\in \mathbb{R}^{d_{\b{u}}}$,
$\b{o}_{t}\in \mathbb{R}^{d_{\b{x}}}$, respectively. Typical artificial neurons
have a bias term. For the sake of notational simplicity and without loss of
generality, this bias term is suppressed here. The bias term can be induced by
considering that one of the coordinates is set to 1 for all times. The dynamics
of our RNN network is as follows:
\begin{eqnarray}
  \b{u}_{t+1}&=& \b{Fu}_{t}+\b{Gx}_{t},\label{eq:u}\\
  \b{o}_{t+1}&=& \b{Hu}_{t+1},\label{eq:o}\\
\hat{\b{x}}_{t+1} &=&\b{o}_{t+1},\label{eq:hat}
\end{eqnarray}
where $\b{F}\in \mathbb{R}^{d_{\b{u}}\times d_{\b{u}}}$, $\b{G}\in
\mathbb{R}^{d_{\b{u}}\times d_{\b{x}}}$, and $\b{H}\in
\mathbb{R}^{d_{\b{x}}\times d_{\b{u}}}$.

In these equations:
\begin{itemize}
    \item
        (\ref{eq:u}) describes the dynamics of the hidden state of the RNN, this is \emph{state equation},
    \item
        (\ref{eq:o}) is the \emph{observation equation}, which provides the mapping of the state equation
         to the external world, and
    \item
        (\ref{eq:hat}) denotes our approximation that the output of the RNN will be used to estimate the input
        in the next time instant. This is the \emph{approximation equation}.
\end{itemize}
In what follows, the linear dynamical system of
Eqs.~(\ref{eq:u})-(\ref{eq:hat}) will be referred to as linear recurrent neural
network (LRNN).

\subsection{Problem domain}
Consider time series $\{\b{x}_{1}, \b{x}_{2}, \dots,\b{x}_{T}\}$. We assume
that the identification of the Elman-type network and the approximation of the
reconstructing hidden states may be improved if both the hidden representation
and the parameters of the Elman network are subject to sparsification.
Optimization via sparsification, that is via the $L_{1}$ norm and the related
$\epsilon$-insensitive norm \cite{Vapnik95Nature} will be examined with respect
to optimization via the quadratic $L_{2}$ norm. Optimization concerns both the
hidden representation, that is $\b{u}_{t}$, $t=1,2, \ldots$ and the weights,
that is matrix $\b{F}$. The linear recurrent network assumption allows for a
joined formalism and shall be tested on different time series.

\subsection{Notations}
Let us introduce the following notations:
\begin{description}
    \item[Letter types:]
        Numbers ($b$), vectors\footnote{Vector means column vector here.} ($\b{b}$), matrices ($\b{B}$) are distinguished by
        the letter types.
    \item[Special vectors and matrices:]
         Certain letters denote particular vectors and matrices. Beyond the already
         introduced notations ($\b{x}_{t}, \b{u}_{t}, \b{o}_{t}, \b{F}, \b{G},$ and $\b{H}$)
          $\b{e}$, $\b{E}$ denote a vector and a matrix with all components equal to 1, respectively;
          $\b{I}$ is the identity matrix; $\b{U}$, $\b{X}$ are matrices with column indices corresponding
          to time and representing the whole hidden and input time series, respectively;
          $\b{R}$ is the $\epsilon$-insensitive multiplier matrix (see also matrix norms).
    \item[Operations on matrices:]
        Matrices are subject to the following operations:
        \begin{itemize}
            \item
                $\b{M}^{T}$ denotes the transposed form of the matrix.
            \item
                $tr(\b{M})$ is the trace.
            \item
                $vec(\b{M})$ is the matrix in column vector form. The columns of the matrix
                follow each other in this vector.
            \item
                $\b{A}\otimes\b{B}$ denotes the Kronecker product of the matrices:
                $\b{A}\otimes\b{B}:=\left[a_{i,j} \b{B}\right]$.
        \end{itemize}
    \item[Matrix relations:] Relations between matrices, such as $>$, $\ge$, and so on, concern all coordinates.
    \item[Norms of matrices:]
        The usual $\epsilon$-insensitive and Euclidean norms can be extended to matrices:
        \begin{itemize}
            \item
                $\left\|\b{M}\right\|_{\b{R}}:=\sum\limits_{i,j}|m_{i,j}|_{{\epsilon}_{i,j}}$,
                   where $|m|_{\epsilon}$ is the \mbox{$\epsilon$-insensitive} cost
                   function, i.e.,
                   \[
                        |m|_{\epsilon}:=\{0, \text{ if } |m|\le \epsilon; |m|-\epsilon,  \text{ otherwise}\},
                  \]
                  and $\b{R}=[\epsilon_{ij}]$.
            \item
            $\left\|\b{M}\right\|_{\b{K}}^{2}:=tr(\b{M}^{T}\b{KM})$  is the norm of the matrix induced
            by positive definite matrix $\b{K}$.
        \end{itemize}
        For the case of vectors, choices of $\b{R}=0$ or $\b{K}=\b{I}$, simplifies to the well known $L_{1}$ and $L_{2}$
        norms, respectively. For the sake of simplicity and when no confusion may arise, $L_{1}$ and $L_{2}$
        norms will be denoted by $||\cdot||_{1}$ and $||\cdot||^{2}$, respectively.
     \item[Linear multi-term expression:]
        Expression of type $M(\b{Z})= \sum\limits_{i=1}^{n}\b{P}_{i}\b{Z}\b{Q}_{i}+\b{B}$ is called linear multi-term expression in matrix
        $\b{Z}$.
 \end{description}
Notation $size(.)$ will be the shorthand to denote the sizes of matrices. For
example, for matrix $\b{Z}\in \mathbb{R}^{a \times b}$, $size(\b{Z},1)=a$ and
$size(\b{Z},2)=b$. Temporal shift is denoted by operator $z$, which increases
all temporal indices by one and produces $\b{x}_{t}\rightarrow \b{x}_{t+1}$
transitions. Operator $z$ may also act on the columns of matrices. For example,
for matrix $\b{X} = [\b{x}_{1},\dots,\b{x}_{T}]$,
$z\b{X}=[\b{x}_{2},\dots,\b{x}_{T+1}]$.

\section{\label{sec:Methods}Formalism and derivations}

There are three constraints that we considered when building the framework:
\begin{enumerate}
    \item The framework is constrained to LRNN dynamics.
    \item The goal of the framework is the approximation and the prediction of time series $x_{t}$.
    \item Regularization, which can ensure unambiguous solution, shall be the tool of sparsification.
\end{enumerate}
In turn, both approximation and regularization need to be considered.

\subsection{Unified description of approximation and regularization}

\begin{figure*}
  \includegraphics[scale=0.5]{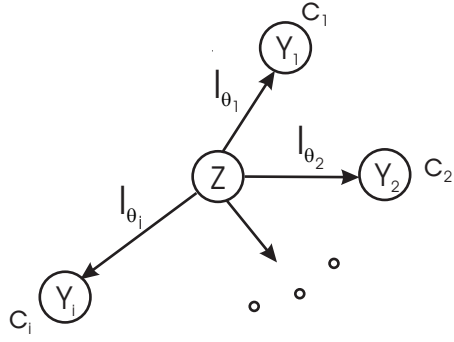}
  \caption{\label{fig:Model}Directed graph of approximation and regularization. Task: estimate unknown source $\b{Z}$
  by assuming parameterized $\theta_{i}$ transforms, i.e., certain $l_{\theta_{i}}(\b{Z})$ forms such that
  the transformed values approximate prescribed values $\b{Y}_{i}$ for all $i = 1, 2, \ldots$. Estimation
  means that the quality of approximation as determined by the cumulated costs of each transformation $c_{i}$
  is minimum.}
\end{figure*}

Let us estimate the value of source $\b{Z}$ subject to the conditions that
different $l_{\theta_{i}}(\b{Z})$, the $\theta_{i}$ ($i=1, 2, \ldots , m$)
transformed values of $\b{Z}$, approximate certain $\b{Y}_{i}$ values and also
minimize certain $c_{i}$ cost functions. Formally the goal is
\begin{equation}
     \sum_{i}c_{i}(\b{Y}_{i}-l_{\theta_{i}}(\b{Z}))\rightarrow\min_{\b{Z}}.
\end{equation}
This goal is illustrated by the directed graph of Fig.~\ref{fig:Model}, which
depicts the costs that may arise. Every edge represents certain transformation,
and every cost is represented by a node. Some nodes are approximation nodes,
i.e., represent approximation costs, whereas others are regularization nodes
and represent costs associated with the regularization.

In our case, $\b{Z}$ corresponds to state sequences and the parameters of the
LRNN system as it will be detailed below.

\subsection{LRNN cost function}
The following cost function fulfills our requirements
\begin{eqnarray}
f(\Theta,\b{U})&:=&\lambda_{state}\left\|\cdot\right\|_{state}+\lambda_{appr}\left\|\cdot\right\|_{appr}+\lambda_{\Theta}
reg(\Theta)+\lambda_{\b{U}} reg(\b{U})\\
&&(\lambda_{state},\lambda_{appr},\lambda_{\Theta}, \lambda_{\b{U}}>0)\nonumber
\end{eqnarray}

Here, $reg(\cdot)$ denotes regularization terms, $\Theta$ represents the parameters of the LRNN system
($\Theta:=\left(\b{F},\b{G},\b{H}\right)$), and $\b{U}$ is the internal state sequence
($\b{U}:=\left[\b{u}_{1},\dots,\b{u}_{T}\right]$) and $\b{Z}$ stands for $\Theta$ or $\b{U}$. Optimizations over
$\Theta$ and $\b{U}$ were iterated repeatedly. This procedure ensures convergence, at least to a local minimum.

Compact formulation can be gained by introducing matrix $\b{X}$, which
represents state sequence and matrices $\b{C}^{b},\b{C}^{e}$, which shall be
useful for describing temporal evolutions:
\begin{itemize}
    \item
        State sequence matrix $\b{X}:=\left[\b{x}_{1},\dots,\b{x}_{T}\right]$.
    \item  Matrix $\b{C}^{b}$ ((cut-the-beginning) performs transformation
        \begin{eqnarray*}
            \b{u}_{1},\dots, \b{u}_{T} &\rightarrow &
            \b{u}_{2},\dots,\b{u}_{T},
        \end{eqnarray*}
        whereas matrix $\b{C}^{e}$ (cut-the-end) acts like
        \begin{eqnarray*}
            \b{u}_{1},\dots, \b{u}_{T} &\rightarrow & \b{u}_{1},\dots,\b{u}_{T-1},\\
            \b{x}_{1},\dots, \b{x}_{T} &\rightarrow & \b{x}_{1},\dots,\b{x}_{T-1}.
        \end{eqnarray*}
        That is
        \begin{equation*}
        \b{C}^{b}:=\left[\begin{array}{ccccc}0&0&\dots&
        0\\1&0&\dots&0\\0&1&0&\dots&\\\vdots&&\ddots&0\\0&\dots&&1\end{array} \right], \b{C}^{e}:=\left[\begin{array}{cccc}1&0&\dots&
        0\\0&1&0&\dots\\\vdots&0&\ddots&0\\0&\dots&&1\\0&0&\dots&0\end{array}\right]\in\mathbb{R}^{T\times(T-1)}.
        \end{equation*}
\end{itemize}
Now, we can write the approximation property as
\begin{equation}
\left\|\b{H}(\b{FU}+\b{GX})-z\b{X}\right\|\approx 0 \label{eq:appr},
\end{equation}
whereas the state equation assumes the following form:
\begin{equation}
 \b{UC}^{b} = (\b{FU}+\b{GX})\b{C}^{e}\label{eq:Mtx-state}.
\end{equation}
Equations~(\ref{eq:appr})-(\ref{eq:Mtx-state}) are of affine conditions either
in $\b{U}$ or in fixed LRNN parameters $\b{F}$, $\b{G}$, and -- in a
degenerated sense -- also in matrix $\b{H}$. Notice that both (\ref{eq:appr})
and (\ref{eq:Mtx-state}) are linear multi-terms in the LRNN parameters and in
the internal state, respectively, provided that the other terms are kept
constant. Thus, cost function $f$ can be iteratively optimized by alternatively
minimizing over $\Theta$ and $\b{U}$. The cost function decreases in every step
and, upon convergence, locally optimal parameter set and internal state
sequence will be reached, provided that the regularization terms can be
managed. Let us consider the $\left\|\b{Z}\right\|_{\b{K}}^{2}$, or
$\left\|\b{Z}\right\|_{\b{R}}$ regularization expressions. Such regularization
terms
\begin{itemize}
    \item
        can be seen as linear multi-terms (of unit length in $\b{Z}$)
    \item
        ensure uniqueness in each iterative step if $\b{K}=\b{I}$ and $\epsilon=0$
        (i.e., in $L_{1}$ norm).
\end{itemize}
The latter condition corresponds to sparsification
\cite{Olshausen97Sparse,Hoyer03Modelling}.

For the sake of simplicity in the form of the equations, we shall restrict our
considerations to $\Theta:=\b{F}$. We shall study costs that emphasize
reconstruction with sparsification

\begin{eqnarray}
f_{s}(\b{F},\b{U})&:=&\lambda_{appr}\left\|\b{H}(\b{FU}+\b{GX})-z\b{X}\right\|_{\b{R}}\nonumber\\
&&+\lambda_{state}\left\|(\b{FU}+\b{GX})\b{C}^{e} -
\b{UC}^{b}\right\|_{\b{R}}+\lambda_{\b{F}}\left\|\b{F}\right\|_{1} +
\lambda_{\b{U}}\left\|\b{U}\right\|_{1}.\label{eq:fs}
\end{eqnarray}

Cost function $f_{s}$ is made of $\epsilon$-insensitive terms ($\b{R}$ with
$\epsilon \ge 0$) of Support Vector Machines (SVMs, see, e.g.,
\cite{Vapnik95Nature,evgeniou00regularization,Herbrich02Learning} and the
references therein). This separation of the $L_{1}$ and $\epsilon$-insensitive
norms is somewhat arbitrary. Both norms are used in order to demonstrate the
diversity enabled by the mathematical framework. Different combinations are, of
course, possible, but no effort was made to find the better combination.

For comparison, we shall study systems equipped with quadratic cost function
$f_{q}$:
\begin{eqnarray}
f_{q}(\b{F},\b{U})&:=&\lambda_{appr}\left\|\b{H}(\b{FU}+\b{GX})-z\b{X}\right\|^{2}\nonumber\\
&&+\lambda_{state}\left\|(\b{FU}+\b{GX})\b{C}^{e} -
\b{UC}^{b}\right\|^{2}+\lambda_{\b{F}}\left\|\b{F}\right\|^{2} +
\lambda_{\b{U}}\left\|\b{U}\right\|^{2}.\label{eq:fq}
\end{eqnarray}

\subsection{Optimization of cost functions}
We shall simplify the description by transcribing cost functions $f_{s}$ and
$f_{q}$ into vectorial form. As it has been noted before, each term --
irrespective of its $\epsilon$-insensitive or quadratic nature -- can be seen
as a linear multi-term in the corresponding iteration step. Thus, it is
sufficient to consider two types of costs:

\begin{eqnarray}
f_{1}(\b{Z})&:=&\lambda\left\|\sum\limits_{i=1}^{n}\b{L}_{i} \b{Z} \b{M}_{i}-\b{N}\right\|_{\b{R}}\rightarrow\min_{\b{Z}},\\
f_{2}(\b{Z})&:=&\lambda\left\|\sum\limits_{i=1}^{n}\b{L}_{i} \b{Z}
\b{M}_{i}-\b{N}\right\|^{2}_{\b{K}}\rightarrow\min_{\b{Z}},
\end{eqnarray}
where $\b{Z}$ denotes the variable subject to optimization in the given step,
i.e. $\b{F}$, or $\b{U}$. Such terms can be added to form the full cost
functions $f_{s}$ and $f_{q}$, respectively. We shall make use of the forms:
\begin{eqnarray}
    vec(\b{BCD})  &=&  (\b{D}^{T}\otimes \b{B})vec(\b{C}),\label{vec:lin0}\\
    tr(\b{B}^{T}\b{C}) & = &vec(\b{B})^{T} vec(\b{C}),\label{tr:lin0}\\
    tr\left(\b{BX}^{T}\b{CYD}\right) & = & vec(\b{X})^{T} \left[\left(\b{B}\otimes
        \b{I}_{size(\b{X},1)}\right)^{T}\left(\b{D}^{T}\otimes\b{C}\right) \right] vec(\b{Y}).\label{tr:kvadr0}
\end{eqnarray}
For the derivation of Eqs.~(\ref{vec:lin0}) and (\ref{tr:lin0}), see
\cite{Minka97Old}. The derivation of Eq.~(\ref{tr:kvadr0}) as well as other
details can be found in the Appendix. We have
\begin{eqnarray*}
\lefteqn{\min_{\b{Z}} f_{1}(\b{Z})\Leftrightarrow}\\
&\Leftrightarrow& \min_{\b{y}}\b{w}^{T}\b{y} \text{, provided that } \{\b{Dy}\le \b{q}\}\text{, where}\\
&&\b{y} := \left[\begin{array}{l}\b{z}\\\b{a}\\ \b{a}^{*}\end{array}\right],
\b{w} := \left[\begin{array}{c}\b{0}\\\lambda \b{e}\\\lambda \b{e}\end{array}\right], \b{M}_{\b{L}}:=\sum\limits_{i=1}^{n}\b{M}_{i}^{T}\otimes \b{L}_{i}, \\
&&\b{D} := \left[\begin{array}{rrr}
\b{M}_{\b{L}}&-\b{I}&\b{0}\\-\b{M}_{\b{L}}&\b{0}&-\b{I}\\\b{0} & -\b{I}
&\b{0}\\\b{0}&\b{0}&-\b{I}\end{array}\right],\b{q} :=
\left[\begin{array}{l}\b{r}+\b{n}\\\b{r}-\b{n}\\
\b{0}\\\b{0}\end{array}\right].
\end{eqnarray*}

\begin{eqnarray*}
f_{2}(\b{Z})&=&\frac 12\b{z}^{T}\b{H}\b{z}+\b{f}^{T}\b{z}\,(+const)\text{,
where }\b{Z}\in\mathbb{R}^{size(\b{Z},1)\times size(\b{Z},2)}.
\end{eqnarray*}
Optimizations can be executed by simply collecting the representative terms,
because all terms are either in linear or in quadratic forms.

\section{\label{sec:Experiments}Experimental studies}
In the numerical studies, advantages of the joined parameter and
representational sparsification -- to be referred to as the \emph{`sparse
case'} -- were explored for the LRNN architecture. The error of the
approximation was monitored as a function of iteration number. The following
test examples were chosen:
\begin{description}
    \item[MG-17,MG-30:]
        The Mackey-Glass time series were originally introduced for modelling the behavior of blood cells
        \cite{Mackey77Oscillation} using delayed difference equations:
                    \[
                      \dot x_{t}=\frac{a x_{t-\tau}}{1+x^{c}_{t-\tau}}-b x_{t}.
                    \]
        This equation is one of the standard benchmark tests. We used the usual parameterizations of the Mackey-Glass
        time series ($a=0.2, b=0.1, c=10, \tau=17$ and $30$) under sampling time equal to $6s$. The time
        series was computed by $4^{th}$ order Runge-Kutta method.
    \item[FIR-Laser:]
            This time series is series `$A$' of the competition at Santa Fe on time series prediction
            \cite{Weigend93Time}. It is the measured intensity pattern of far infrared (FIR) laser
            in its chaotic state. The time series will be referred to as FIR-Laser series.
    \item[Henon:] The dynamical model
        \begin{eqnarray*}
            x_{t+1} &=& 1 - a x_{t}^{2} + y_{t},\\
            y_{t+1} &=& b x_{t}
        \end{eqnarray*}
        was proposed by Henon \cite{Henon76Two}. We use standard parameters ($a=1.4$, $b=0.3$),
        generate the Henon-attractor from pair $(x_{t},x_{t-1})$, and predict $x_{t}$.
\end{description}

We shall approximate one dimensional series ($d_{\b{x}}=1$) of length $T$ with
LRNN. 10 different $T$ values ($T= 10, 20, \ldots 100$) shall be used. Matrix
$\b{H}$, which approximates the output from the hidden activites of the LRNN,
is a projection to the first coordinate.

Observation, that is matrix $\b{H}$ of the LRNN, shall be the projection to the
first
 coordinate Each coordinates of matrices $\b{F},\b{G}$ were initialized
 by drawing them independently from the uniform distribution over $[0,1]$.
 $\epsilon$-insensitivity was chosen as 0.05, that is, $\b{R}=0.05\b{E}$. The dimension of the hidden state was
 set to $d_{u}=4$.

Normalization steps were applied to allow quantitative comparisons:
\begin{itemize}
    \item
        Time series were scaled to vary between $[-1,1]$.
    \item
        The $\lambda$'s of Eqs.~(\ref{eq:fs}) and (\ref{eq:fq}) were chosen
        to allow for similar contributions  for every term, independently from the dimension
        of the hidden state of the RNN and the number of the adjustable parameters:
        \[
            \lambda_{appr}=\frac{1}{T d_{\b{x}}}, \quad\lambda_{state}=\frac{1}{(T-1) d_{\b{u}}},\quad
            \lambda_{\b{U}}=\frac{1}{T d_{\b{u}}},
            \quad\lambda_{\b{F}}=\frac{1}{d_{\b{u}}^{2}}\text{ .}
        \]
\end{itemize}

Optimizations for the sparse case [see Eq.~(\ref{eq:fs})] were compared with
optimizations using quadratic cost function [Eq.~(\ref{eq:fq})]. There is a
large diversity of possible mixed cost functions. No effort was made to select
the main contributing terms to the differences in our results. Such differences
might change from problem to problem.

\subsection{\label{sec:Results}Results}

Optimization warrant locally optimal solutions. To overcome this problem, in
each particular experiment, 20 random initializations of matrices $\b{F}$ and
$\b{G}$ were made and 20 different optimizations were executed. Averages of the
costs $f_{q}$ and $f_{s}$ were computed and shall be depicted.

\begin{figure*}[h!]
  \includegraphics{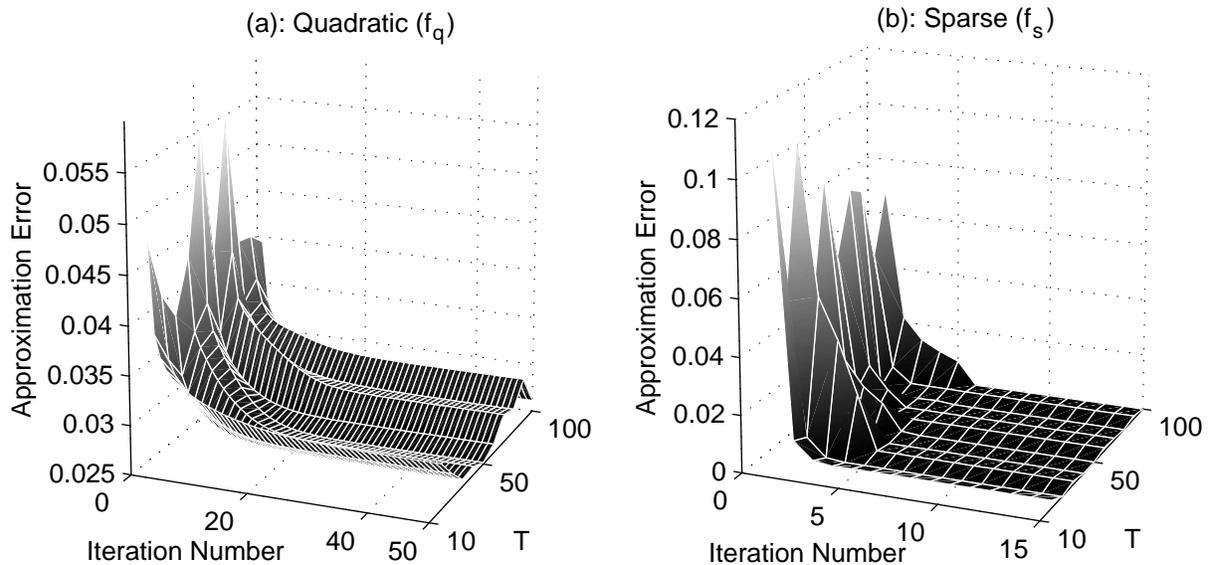}
  \caption{\label{fig:ApprIt}Performance under quadratic cost and for the sparse case.
  Approximation error versus iteration number and the length of the time series ($T$) approximated for the
  case of the MG-30 problem. Results are averaged over 20 random experiments. (a): ($f_{q}$); the case for quadratic cost
  function. (b): ($f_{s}$); the sparse case. Convergence speed of iteration is faster for the sparse case -- note
  the differences in scales.}
\end{figure*}

Approximation error as a function of iteration number is shown in
Fig.~\ref{fig:ApprIt} for the case of the MG-30 problem. Results are averaged
over 20 experiments. Similar results were found in the other experiments, too.
Time series of different length were tried. According to the figure,
sparsification has its advantages, it makes convergence faster and more
uniform:

The approximation produced small errors for the quadratic norm up the lengths
about 50. The iteration number required for approximate convergence (the
`knees' of the curves) was below 10. For time series having lengths between 50
and 100, the error increased considerably and the typical iteration number
required for approximate convergence increased to 20. One may say that up to
100 step long time series and for the quadratic case, 20 iterations are
satisfactory to reach stable approximation error, i.e., the close neighborhood
of the local minimum.

For the sparse case, the situation is different. Convergence is very fast, 6
iterations were satisfactory to reach stable approximation error in all cases.
Errors converged in a few steps.

\begin{figure*}[h!]
    \includegraphics{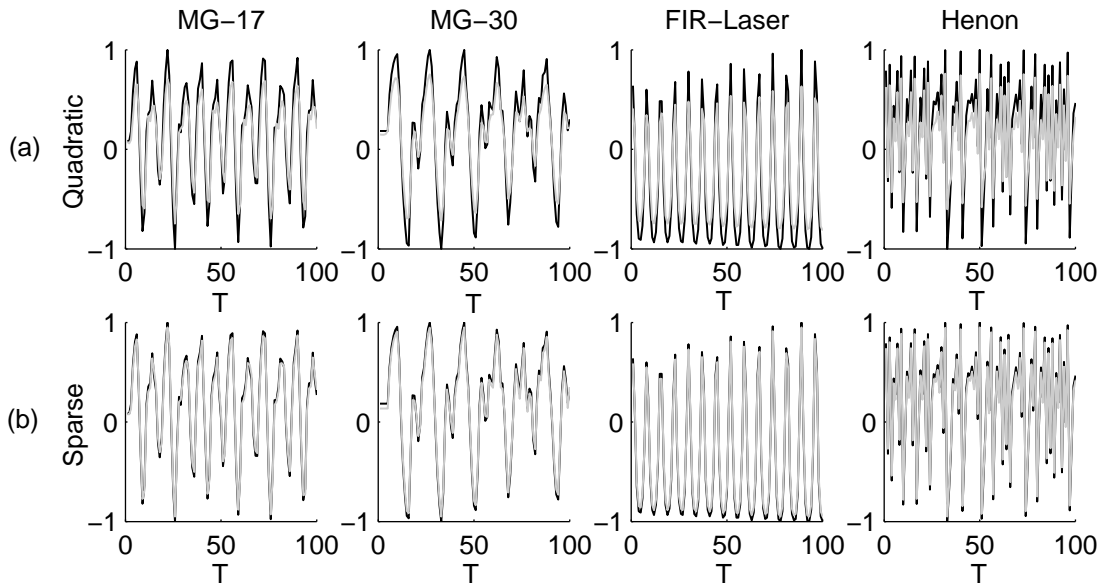}
    \caption{\label{fig:ApprPL}Predictions of LRNN systems trained by different methods and for different
    problems. Problem set:  MG-17, MG-30, FIR-Laser, Henon systems are shown in columns 1, 2, 3, and 4,
    respectively. (a): `Quadratic', prediction for quadratic cost function. (b): `Sparse', prediction
    for the sparse case.  Length of approximated time series: 100.
    Black: original time series. Gray: approximation. For detailed results, see, Fig.~\ref{fig:ApprT}.
}
\end{figure*}

\begin{figure}[h!]
    \includegraphics[width=16cm]{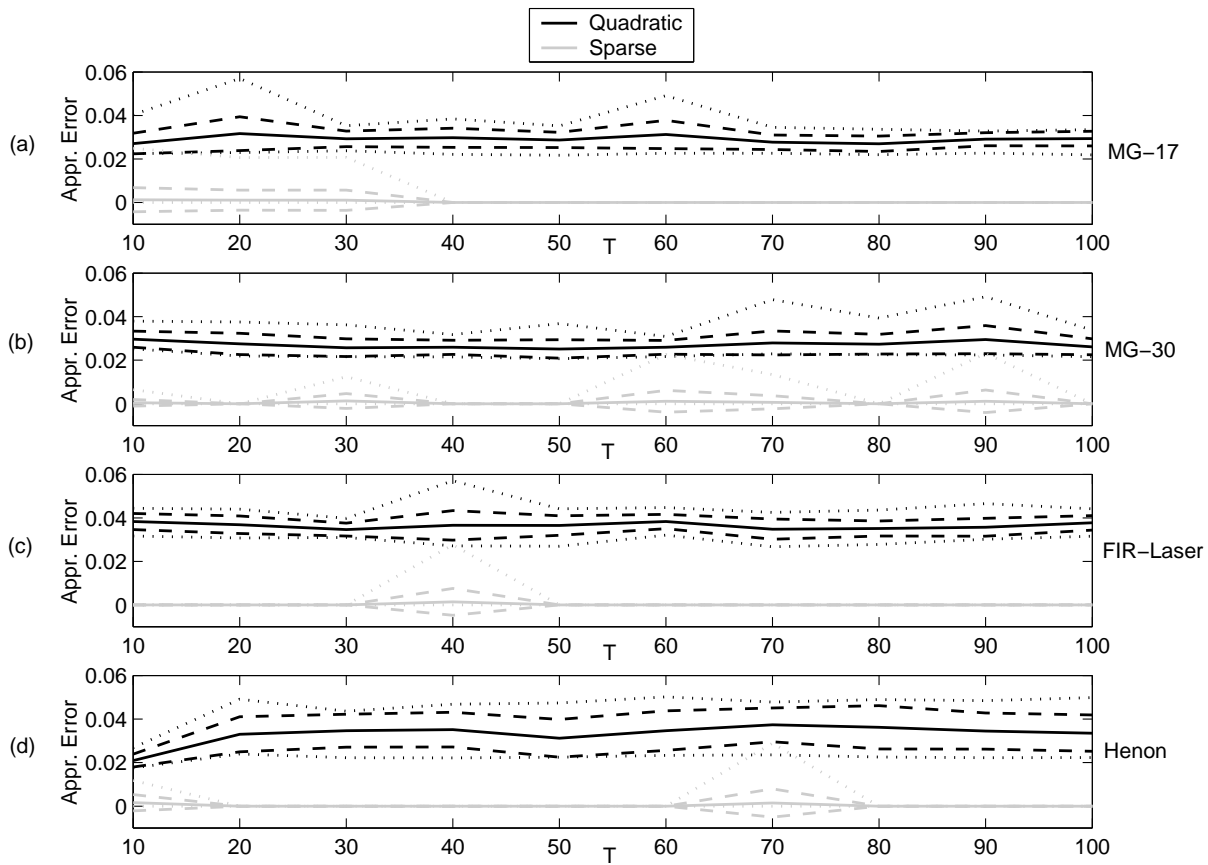}
    \caption{\label{fig:ApprT}Prediction error averaged over 20 experiments. Experiments are averaged over 20 randomly initialized and optimized LRNN models for the 4
     time series examples ((a): MG-17, (b): MG-30, (c): FIR-Laser, (d): Henon). Horizontal axis: time $T$. Vertical axis: error.
     Solid line: averaged magnitude. Dashed line: standard deviation. Dotted line: best and worst cases.
     Black: results for quadratic cost function. Gray: results for the sparse case.
     Prediction error averaged over $T$ is on the order of $10^{-4}$ for
     the gray curves. See also Table~\ref{tab:epsError}.}
\end{figure}

Predictions of the optimized LRNNs are shown in Fig.~\ref{fig:ApprPL} for 100
time steps. Sparsification again shows improved interpolating capabilities.
This impression is reinforced  by Fig.~\ref{fig:ApprT}, which depicts averages
of 20 randomly initialized computations, alike to the one shown in
Fig.~\ref{fig:ApprPL}. Averaged prediction errors, that is
$\lambda_{appr}\left\|\b{H}(\b{FU}+\b{GX})-z\b{X}\right\|^{2}$ and
$\lambda_{appr}\left\|\b{H}(\b{FU}+\b{GX})-z\b{X}\right\|_{\b{R}}$ values for
the quadratic and for the sparse case are depicted in the figure. Respective
standard deviations as well as the best and the worst cases are also provided.

Table~\ref{tab:epsError} shows small errors for the $\epsilon$-insensitive
norm.
       \begin{table}[h]
              \caption{\label{tab:epsError}Time averaged prediction error for the $\epsilon$-insensitive cost.}
              \begin{tabular}{|c|c|c|c|}
                \hline
                MG-17 & MG-30 & FIR-Laser & Henon \\
                \hline
                $3.3\cdot 10^{-4}$ & $4.6\cdot 10^{-4}$ & $1.3\cdot 10^{-4}$ & $3\cdot 10^{-4}$\\
                \hline
              \end{tabular}
       \end{table}
According to Fig.~\ref{fig:ApprT}, the deviation from the average is also small
for long intervals, but -- due to the nature of this cost function, which
measures costs relative to 0.05  -- sometimes it is relatively large. The
comparison of the different costs is somewhat arbitrary. For time series
generated by MG-17, MG-30, FIR-Laser, and Henon systems, the LRNN trained by
joined parameter and representational sparsification seems to exhibit a factor
of three better approximation than the LRNN using the quadratic norm. Such
crude estimation can be gained by comparing 0.158, (the square root of 0.025,
which is an approximate low value for the error of the quadratic norm
(Fig.~\ref{fig:ApprIt})) with the value of $\epsilon =0.05$, the estimation for
the sparse case from above. We have compared the normalized root mean square
error for the two norms: If the measured value is $o_{t}$ and the target is
$x_{t}$ then
    \[
        NMRSE:=\frac{\sqrt{\langle(o_{t}-x_{t})^{2}\rangle}}{\sqrt{\langle(x_{t}-\langle
        x_{t}\rangle)^{2}\rangle}},
    \]
where $\langle\cdot\rangle$ denotes averaging over $t$. For the quadratic and
the $\epsilon$-insensitive cases we received roughly $NRMSE=0.3$ and
$NRMSE=0.08$, respectively. That is, comparison using NRMSE, which favors the
quadratic norm, there is an advantage of a factor 3 for the
$\epsilon$-insensitive norm.

Finally, we note that long term prediction is hard for the linear approach:
Apart from a zero measure subset, linear recurrent networks either converge or
diverge exponentially.

\section{\label{sec:Discussion}Discussions}

First, let us mention that convergence of certain sparsification methods can be questioned \cite{olshausen96learning}.
The method we presented makes use of the $L_1$ norm for sparsification, warrants convergence and is an extension of
previous architectures, which did not consider a hidden predictive model.

We should address the following questions:
\begin{enumerate}
    \item\label{1} What is the advantage of sparse representation?
    \item\label{2} What is the advantage of weight sparsification, or pruning?
\end{enumerate}

Question~\ref{1} is intriguing, because neuronal coding seem to apply
sparsification as the basic strategy. Neuronal firing in the brain is sparse,
which could serve the minimization of energy consumption
\cite{baddeley96efficient}. It is now well established that joint constraints
of sparsification and reconstruction (i.e., generative) capabilities produce
receptive fields similar to those found in the primary visual cortex
\cite{Olshausen96Emergence}. More sophisticated architectures, which also apply
sparsification can reproduce more details of the receptive fields
\cite{rao99predictive}.

Are there other advantages beyond energy minimization? Recently, independent
component analysis (ICA)
\cite{jutten91blind,comon94independent,bell95information}, an information
transfer optimizing scheme, has been related to sparsification
\cite{hyvarinen99sparse2}. It has been shown that ICA and appropriate
thresholding corresponds to well established denoising methods. Denoising, that
is the removal of structureless high entropy portion of the input and the
uncovering of structures are important tasks for learning. Subliminal
statistical learning is indeed, a strategy applied by the brain
\cite{watanabe01perceptual}. ICA with denoising capabilities, called sparse
code shrinkage (SCS) \cite{hyvarinen99sparse2}, enables \emph{local} noise
estimation and \emph{local} noise filtering. SCS is most desirable, because it
optimizes noise filtering with respect to the experienced inputs.

Another advantage emerges if sparsified representation is embedded into
reconstruction networks. This construct can produce overcomplete
representations \cite{Olshausen96Emergence} which, by construction, exhibit
graceful degradation. Also, overcomplete representations can produce very
different outputs for similar inputs, a most desirable feature for
categorization and decision making.

It is easy to see that norm $L_{1}$ biases the system towards sparse
representation compared to the $L_{2}$ norm: The $L_{1}$ norm produces larger
costs for small signals and depresses small signals more efficiently. This
statement seems more general, because the $\epsilon$-insensitive cost function
is closely related to sparsification networks \cite{girosi98anequiv}.

There are different noise models behind the $L_{1}$ and the $L_{2}$ norms. In
such considerations, the norm of the approximation is seen as the
log-likelihood of the distribution of the noise. The $L_{1}$ and the $L_{2}$
norms, as well as that of the $epsilon$-insensitive norm are all incorporated
into the following log-likelihood expression \cite{pontil00onthenoise}:
\begin{equation}
 V(x) = - \log \int_{0}^{\infty} d\beta \int_{-\infty}^{\infty} dt \sqrt{\beta} e^{-\beta (x-t)^{2}} P(\beta,t),\label{eq:losses}
\end{equation}
where $x$ is the argument of the respective norms. For example, quadratic loss
function emerges if $P(\beta,t)=\delta(\beta-\frac{1}{2\sigma^{2}})\delta(t)$,
where $\delta(.)$ is Dirac's delta function. The $L_{1}$ loss is recovered by
setting $P(\beta,t)=\beta^{2} \exp (-\frac{1}{4\beta^{2}})\delta(t)$. The
expression for $\epsilon$-insensitive loss in a factorized form for
$P(\beta,t)$ is as follows: $P(\beta,t)=\frac{C}{\beta^{2}} \exp
(-\frac{1}{4\beta^{2}})\lambda_{\epsilon}(t)$, where
$\lambda_{\epsilon}(t)=\frac{1}{2(\epsilon+1)}\left(
\chi_{[-\epsilon,\epsilon]}(t)+\delta(t-\epsilon)+\delta(t+\epsilon) \right)$,
$\chi_{[-\epsilon,\epsilon]}$ is the characteristic function of the interval
$[-\epsilon,\epsilon]$, and C is a normalization constant. The interpretation
is that the noise affecting the data is additive and Gaussian, but its mean can
be different from zero. The variance and mean of the noise are random variables
with given probability distributions. For diminishing range of the non-zero
mean Gaussian distributions, the expression approximates the $L_{1}$ norm
\cite{pontil00onthenoise}. In turn, both the $L_{1}$ norm and the
$\epsilon$-insensitive norm correspond to super-Gaussian noise distributions.

Concerning Question~\ref{2}, sparsification of weight matrices, e.g.,
exponential forgetting, pruning of weights, as well as other computational
means have been studied over the years. The interested reader is referred to
the literature on this broad field \cite{haykin99}. Question~\ref{2} seems to
have a particular advantage, called the Occam's razor principle. In reverse
engineering, such as the search for the (hidden) parameters of an LRNN, there
are many solutions having identical or similar properties. All of these could
be the \emph{the solution} of the task. Which one is the right one? Or, at
least, which one is the most probable? The answer of the philosopher is that
the simplest \emph{explanation}, that is, the one that makes the least
assumptions is the best, or the most probable. Similar ideas have emerged in
information theory, within the context of Kolmogorov complexity
\cite{cover91elements}. A bridge has been built to connect the idea of
\emph{least number of assumptions} and concepts of complexity. This is the
minimum description length principle \cite{rissanen78mdl}, which makes use of
complexity measures both for the data and for the assumed family of probability
distributions, that is for the model. The best, i.e., the most probable
description, has the minimum complexity. Important connections between the
minimum description length principle and regularization theory have been worked
out in the literature \cite{mackay92phd,hinton94autoencoders}. For a recent
review on this subject, see \cite{chen02different}. One may think of weight
sparsification as a method, which decreases the number of parameters and thus
searches for simpler descriptions, provided that the precision of the
parameters is bounded.

Our results indicate that local minima are better distributed for $L1$ and/or
$\epsilon$-insensitive norms than for the $L_{2}$ norm within the problem set
we studied. However, it is known that there should be other problem sets, where
the quadratic cost is superior to the absolute value \cite{wolpert97nofree}. It
is easy to create an example: quadratic cost on the parameters suits parameter
sets, which exhibit Gaussian distribution. In turn, the success of the $L_{1}$
and $\epsilon$-insensitive norms raises the question about the type of
problems, which profit from this cost function family.

We need to examine the statistical properties of phenomena in nature. It seems
that these phenomena have special distributions. The distributions are
super-Gaussian, sometimes exponential and sometimes they have very heavy tails,
e.g., they can be characterized by power-law distribution in a broad range.
This looks typical for natural forms (c.f. fractals) and evolving/developing
systems (see, e.g., \cite{albert02statistical} and references therein).

We have measured the four time series: Mackey-Glasses (MG-17, MG-30),
FIR-Laser, Henon. The distribution of distances between zero crossings of time
series of 10,000 step long were analyzed.
\begin{enumerate}
    \item
        The kurtosis of the experienced distributions were computed, according to
        \[
            kurtosis(X)=\frac{E\left[X-\mu\right]^{4}}{\sigma^{4}}-3,
        \]
        where $E$ denotes expectation value, $\mu$ is the mean of variable $X$, $\sigma$ is its standard deviation.
        Positive kurtosis is typical for super-Gaussian distributions, which have heavy tails. Results are shown in
        Table~\ref{tab:kurt}.
        \begin{table}[h]
          \caption{\label{tab:kurt}Kurtosis of distances between zero-crossings}
          \begin{tabular}{|c|c|c|c|}
            \hline
            MG-17 & MG-30 & FIR-Laser & Henon \\
            \hline
            -0.92 & 1.11 & 101.8 & 2.89\\
            \hline
          \end{tabular}
        \end{table}
        For all cases, but the MG-17 series, positive kurtosis was found. It is known, however, that series
        MG-17 is at the borderline of chaos: the Mackey-Glass time series becomes chaotic if $\tau>16.8$.
    \item
        The second study concerned the power law behavior. It was found that the zero crossings for the Henon and
        the MG-30 systems approximate power-law distributions, the MG-17 series does not, and the FIR laser is
        in between: it exhibits an erratic descending curve on log-log scale. The slope of the log-log fits were smaller
        smaller than $-1$, except for the MG-17 time series.
\end{enumerate}
The problems we studied, typically have heavy tailed distributions. In turn,
the $L_{1}$ and/or the $\epsilon$-insensitive norms suit such problems better
than $L_{2}$ norm, because the $L_{1}$ norm corresponds to the Laplace (i.e.,
double sided exponential) distribution, which has a heavier tail than the
Gaussian distribution. Power-law distributions, however, are even more at the
extremes in this respect. Because such heavy tailed distributions are typical
in nature (see, e.g., \cite{bak96how,Holland98Emergence}), one may find even
better norms when identification of emergent behaviors is the issue.

\section{\label{sec:Conclusion}Conclusion}
We have put forth a joint formalism for sparsification of both the
representation and the structure of neural networks. The approximation was
applied to a recurrent neural network and the joined effect was studied.
Experiments were conducted on a number of benchmark problems, such as the
Mackey-Glass with parameters 17--30, the emission of a far-infrared laser, and
the Henon time series. Our results suggest that the joined constraint on
sparsification is advantageous on these examples. Namely, optimizations were
faster and had local optima exhibited better interpolating properties for the
sparsified case than for traditional quadratic norm. We have argued that these
findings are due to the underlying statistics of these phenomena: quadratic
norms assume Gaussian noise and distributions with heavy-tails are be better
approximated by norms with lower degrees. Joint advantages of the removal of
structureless high entropy, a feature of representational sparsification, and
the discovery of structures with sparse networks needs to be further explored.

\appendix
\section{Derivation of Eq.~(\ref{tr:kvadr0})}
\begin{eqnarray*}
        tr\left(\b{BX}^{T}\b{CYD}\right)&=&tr\left[\left(\b{XB}^{T}\right)^{T}
        \left(\b{CYD}\right)\right]\\&=& vec\left(\b{IXB}^{T}\right)^{T} vec\left(\b{CYD}\right)\\
        &=&\left[\left(\b{B}\otimes \b{I}\right)vec\left(\b{X}\right)\right]^{T}\left[\left(\b{D}^{T}\otimes\b{C}\right)vec(\b{Y})\right]\\
        &=&vec(\b{X})^{T} \left[\left(\b{B}\otimes \b{I}_{size(\b{X},1)}\right)^{T}\left(\b{D}^{T}\otimes\b{C}\right) \right] vec(\b{Y}).
\end{eqnarray*}

\section{Computing the norms}
\begin{eqnarray*}
\lefteqn{\min_{\b{Z}} f_{1}(\b{Z})\Leftrightarrow}\\
&\Leftrightarrow&\min_{\{\b{Z},\b{A}^{(*)}\}}\left\{\lambda
tr\left[\b{E}^{T}\left(\b{A}+\b{A}^{*}\right)\right]\right\},\\
&&\text{provided that }\left\{
\begin{array}{rll}
\sum\limits_{i=1}^{n}\b{L}_{i} \b{Z} \b{M}_{i}-\b{N}& \le & \b{R} + \b{A}\\
-\left(\sum\limits_{i=1}^{n}\b{L}_{i} \b{Z} \b{M}_{i}-\b{N}\right)& \le & \b{R} + \b{A}^{*}\\
\b{0}& \le &\b{A}^{(*)}\\
\end{array}
\right\}.\\
&\Leftrightarrow&\min_{\{\b{Z},\b{A}^{(*)}\}}\left[\lambda
vec(\b{E})^{T}(vec(\b{A})+vec(\b{A}^{*})\right]=:\left[\lambda\b{e}^{T}(\b{a}+\b{a}^{*})\right],\\
&&\text{provided that } \left\{
\begin{array}{rlll}
\sum\limits_{i=1}^{n}\left(\b{M}_{i}^{T}\otimes \b{L}_{i}\right) \b{z}&-\b{a}& \le & \b{r} + \b{n}\\
-\sum\limits_{i=1}^{n}\left(\b{M}_{i}^{T}\otimes \b{L}_{i}\right) \b{z}&-\b{a}^{*}& \le & \b{r} - \b{n}\\
&\b{0}& \le &\b{a}^{(*)}\\
\end{array}
\right\}.\\
&\Leftrightarrow& \min_{\b{y}}\b{w}^{T}\b{y} \text{, provided that } \{\b{Dy}\le \b{q}\}\text{, where}\\
&&\b{y} := \left[\begin{array}{l}\b{z}\\\b{a}\\ \b{a}^{*}\end{array}\right],
\b{w} := \left[\begin{array}{c}\b{0}\\\lambda \b{e}\\\lambda \b{e}\end{array}\right], \b{M}_{\b{L}}:=\sum\limits_{i=1}^{n}\b{M}_{i}^{T}\otimes \b{L}_{i}, \\
&&\b{D} := \left[\begin{array}{rrr}
\b{M}_{\b{L}}&-\b{I}&\b{0}\\-\b{M}_{\b{L}}&\b{0}&-\b{I}\\\b{0} & -\b{I}
&\b{0}\\\b{0}&\b{0}&-\b{I}\end{array}\right],\b{q} :=
\left[\begin{array}{l}\b{r}+\b{n}\\\b{r}-\b{n}\\
\b{0}\\\b{0}\end{array}\right].
\end{eqnarray*}

\begin{widetext}
\begin{eqnarray*}
f_{2}(\b{Z})&=&\lambda tr\left[\left(\sum\limits_{i=1}^{n}\b{L}_{i} \b{Z}
\b{M}_{i}-\b{N}\right)^{T}\b{K}\left(\sum\limits_{j=1}^{n}\b{L}_{j} \b{Z} \b{M}_{j}-\b{N}\right)\right]\\
 &=&\lambda tr\left[\sum\limits_{i,j}\b{M}_{i}^{T}\b{Z}^{T}\left(\b{L}_{i}^{T}\b{K}\b{L}_{j}\right)\b{Z}\b{M}_{j}-2\sum\limits_{j}
 \b{N}^{T}\b{K}\b{L}_{j}\b{Z}\b{M}_{j}+\b{N}^{T}\b{K}\b{N}\right]\\
 &=&\frac{1}{2} vec(\b{Z})^{T}\left(2\lambda\sum\limits_{i,j}\left\{\left(\b{M}_{i}^{T}\otimes\b{I}_{size\left(\b{Z},1\right)}\right)^{T}
 \left[\b{M}_{j}^{T}\otimes\left(\b{L}_{i}^{T}\b{K}\b{L}_{j}\right)\right]\right\}\right)vec(\b{Z})\\
 &&+\left[-2\lambda\sum\limits_{j} vec(\b{L}_{j}^{T}\b{K}\b{N}\b{M}_{j}^{T})\right]^{T} vec(\b{Z})+\lambda
 tr\left(\b{N}^{T}\b{K}\b{N}\right)\\
 &=:&\frac 12\b{z}^{T}\b{H}\b{z}+\b{f}^{T}\b{z}\,(+const)\text{, where }\b{Z}\in\mathbb{R}^{size(\b{Z},1)\times size(\b{Z},2)}.
\end{eqnarray*}
\end{widetext}

Optimizations can be executed by simply collecting the representative terms,
because all terms are either in linear or in quadratic forms.

\end{document}